\documentclass{article}

\usepackage{siunitx}
\usepackage{multirow}

\PassOptionsToPackage{numbers, comma, square}{natbib}

\usepackage[final]{neurips_2025}

\usepackage[utf8]{inputenc} % allow utf-8 input
\usepackage[T1]{fontenc}    % use 8-bit T1 fonts
\usepackage[hidelinks]{hyperref}       % hyperlinks
\usepackage{url}            % simple URL typesetting
\usepackage{booktabs}       % professional-quality tables
\usepackage{amsfonts}       % blackboard math symbols
\usepackage{nicefrac}       % compact symbols for 1/2, etc.
\usepackage{microtype}      % microtypography
\usepackage{xcolor}         % colors
\usepackage{amsmath}
\usepackage{amsthm}
\usepackage{graphicx}

\newtheorem*{theorem*}{Theorem}

\newtheorem*{lemma*}{Lemma}

\newtheorem*{proposition*}{Proposition}

\title{Scaling Laws and Pathologies of Single-Layer PINNs: Network Width and PDE Nonlinearity}

\author{
  Faris Chaudhry \\
  Department of Computer Science\\
  Imperial College London \\
  \texttt{faris.chaudhry22@imperial.ac.uk} \\
}

\begin{document}

\maketitle

\begin{abstract}
We establish empirical scaling laws for Single-Layer Physics-Informed Neural Networks on canonical nonlinear PDEs. We identify a dual optimization failure: (i) a baseline pathology, where the solution error fails to decrease with network width, even at fixed nonlinearity, falling short of theoretical approximation bounds, and (ii) a compounding pathology, where this failure is exacerbated by nonlinearity. We provide quantitative evidence that a simple separable power law is insufficient, and that the scaling behavior is governed by a more complex, non-separable relationship. This failure is consistent with the concept of spectral bias, where networks struggle to learn the high-frequency solution components that intensify with nonlinearity. We show that optimization, not approximation capacity, is the primary bottleneck, and propose a methodology to empirically measure these complex scaling effects.~\footnote{\url{https://github.com/farischaudhry/pinn-width-vs-nonlinearity}}
\end{abstract}

\section{Introduction}

Physics-Informed Neural Networks (PINNs)~\cite{raissi2019physics} offer a compelling mesh-free paradigm for numerically solving partial differential equations (PDEs) by embedding the governing physics directly into the neural network's loss function. Inspired by the impact of scaling law studies in foundational language and vision models,~\cite{kaplan2020scaling} which have provided an empirical understanding of model performance, a similar quantitative framework for PINNs is critically underdeveloped. The relationship between model capacity, problem complexity, and solution accuracy remains largely unquantified.

At a theoretical level, the Universal Approximation Theorem (UAT)~\cite{cybenko1989approximation, hornik1989multilayer} guarantees that even a Single-Layer Network (SLN) possesses the necessary expressive power to approximate continuous functions. Furthermore, theoretical bounds for function classes like Barron spaces~\cite{barron1993universal} suggest error should decrease with network width $N$ as $\mathcal{O}(N^{-1/2})$, which will correspond to a theoretical scaling exponent of $\alpha = 0.5$ in our experiments. These theorems guarantee existence, but not that gradient-based optimization will find such an approximation in practice. This creates a gap between theoretical expressivity and practical performance, where the primary bottleneck might shift from approximation capacity to optimization challenges in the complex, non-convex loss landscapes of PINNs.~\cite{krishnapriyan2021characterizing} A key mechanism behind this optimization challenge is spectral bias, the tendency for gradient-based optimization to fit low-frequency components of a function far more rapidly and robustly than high-frequency ones.~\cite{wang2022understanding, rahaman2019spectral} As the nonlinearity and complexity of the PDE solution increase, requiring the network to learn these high-frequency features, spectral bias can lead to a failure of the training process.

This work investigates this theory-practice gap. We frame our investigation around the separable scaling law $\text{error} \approx A \cdot N^{-\alpha} \cdot \kappa^{\gamma}$, where $\kappa$ will be our problem-dependent complexity parameter, $\alpha$ is the width scaling exponent, and $\gamma$ is the nonlinearity scaling exponent. This is motivated by the theoretical laws derived from approximation theory. We use SLNs to isolate the effect of width and directly test the limits of the UAT's practical relevance. The goal is to test a two-part hypothesis: (i) that practical SL-PINN training exhibits a baseline optimization pathology where $\alpha \neq 0.5$, and (ii) that this problem is compounded by a second pathology where the separable power law breaks down. We hypothesize that the scaling exponent $\alpha$ is itself a function of the nonlinearity $\kappa$, a non-separable relationship that points to a more fundamental optimization challenge. Our primary contribution is the systematic measurement and analysis of these exponents across a diverse suite of canonical PDEs representing different classes of nonlinear phenomena and PDE types: dispersive (Korteweg-de Vries), transcendental/hyperbolic (Sine-Gordon), and reactive/parabolic (Allen-Cahn).

\section{Related Work}

The challenge of training PINNs for complex physical systems is a topic of intense research. Early work by~\citet{krishnapriyan2021characterizing} provided a foundational characterization of PINN failure modes, arguing that, for stiff or complex PDEs, the bottleneck is not network expressivity but rather a failure of optimization. Our work aims to quantify this failure with a precise scaling law. Other work has explored architectural solutions to handle problem complexity, such as domain decomposition methods like Finite-Basis PINNs, which scale PINNs to larger domains by avoiding single, monolithic networks.~\cite{moseley2023finite} Our study, in contrast, investigates the fundamental scaling properties of the monolithic PINN architecture itself.

This focus on network width reveals a distinction from mainstream deep learning. In many applications, wider networks are understood to facilitate optimization, for example by enabling stable large-batch training.~\cite{you2018effect} Our finding of ``pathological scaling,'' where wider is worse, runs directly counter to this heuristic. This suggests that the PINN optimization landscape has fundamentally different properties, a notion supported by recent theoretical work.~\citet{wang2022understanding} identified gradient flow pathologies and spectral bias as mechanisms that hinder PINN training, especially for high-frequency or multi-scale problems. Further, while the Neural Tangent Kernel (NTK)~\cite{jacot2018neural} provides insight into infinitely-wide networks,~\citet{bonfanti2023challenges} have shown that standard NTK theory is misleading for nonlinear PDEs.

This optimization challenge is particularly acute for fluid dynamics simulations governed by the Navier-Stokes equations. Several studies have used the Reynolds number ($Re$), which quantifies the ratio of inertial to viscous forces, as a natural ``hardness parameter.'' For instance,~\citet{reyes2021learning} observed that PINN performance degraded significantly as $Re$ increased in simulations of flow past a cylinder, identifying this as a key limitation. Similarly,~\citet{arzani2021uncovering} noted that PINNs struggled to capture turbulent flow features at high $Re$. These works highlight a specific instance of the phenomenon we study, but they do not formalize it as a general scaling law. Our hardness parameter $\kappa$ can be seen as a generalization of parameters like $Re$ to other classes of nonlinear PDEs.

\section{Methodology}

The total loss, a weighted sum of the mean squared residuals for the PDE, boundary conditions (BCs), and initial conditions (ICs), is minimized:
\begin{equation}
    \mathcal{L}_{\text{total}} = w_{\text{pde}} \mathcal{L}_{\text{pde}} + w_{\text{bc}} \mathcal{L}_{\text{bc}} + w_{\text{ic}} \mathcal{L}_{\text{ic}}.
\end{equation}
For all models, we use equal weights ($w=1$) and train for 25,000 epochs using the Adam optimizer with a learning rate of $10^{-3}$. We use Adam as it is broadly representative of first-order methods commonly used in the wider deep learning community, allowing us to focus specifically on the architectural effects of width.

\subsection{PDE Suite and Hardness Parameter}

We analyze a suite of canonical PDEs with scalar outputs in one spatial dimension ($x$) and an optional time dimension ($t$). For each nonlinear PDE, we define a tunable hardness parameter $\kappa$ that controls the strength of the nonlinear effects.

\begin{itemize}
    \item \textbf{Poisson Eq.:} $-u_{xx} = \sin(\pi x)$ on $x \in [0,1]$ with Dirichlet BCs. This serves to validate our framework against known theoretical scaling rates for PINNs. Here, there is no dependence on $\kappa$.
    \item \textbf{KdV Eq. (Dispersive):} $u_t + \kappa u u_x + u_{xxx} = 0$. The hardness $\kappa=A$ is the soliton amplitude, which dictates its speed and sharpness.
    \item \textbf{Sine-Gordon Eq. (Hyperbolic/Transcendental):} $u_{tt} - u_{xx} + \kappa \sin(u) = 0$. The hardness $\kappa$ scales the strength of the nonlinear potential term. The IC includes a constraint on $u_t(0,x)$.
    \item \textbf{Allen-Cahn Eq. (Reactive/Parabolic):} $u_t - D u_{xx} + (u^3 - u) = 0$. We define the hardness $\kappa = 1/D$, where smaller diffusion $D$ leads to sharper, harder-to-resolve interfaces.
\end{itemize}

In all three nonlinear cases, this hardness parameter $\kappa$ is designed to control the strength of nonlinear effects, which manifests as an increase in the high-frequency components of the ground truth solution (e.g., sharper solitons or interfaces), thereby creating a direct challenge to the network's inherent spectral bias.

\subsection{Experimental Sweep and Evaluation}

For each PDE, we perform a systematic sweep over:
\begin{itemize}
    \item \textbf{Network Widths $N \in \{16, 32, 64, 128, 256, 512, 1024\}$}.
    \item \textbf{Hardness Parameter $\kappa$}: 7 logarithmically-spaced values for each nonlinear PDE.
    \item \textbf{Activations}: We test both \texttt{tanh} and \texttt{ReLU} activations.
    \item \textbf{Random Seeds}: Each configuration is run with 5 random seeds for statistical robustness.
\end{itemize}
Collocation points are sampled using a Sobol sequence for the Poisson equation and uniform random sampling for the nonlinear PDEs. Error is measured as the mean relative $L_2$ error, $\|\hat{u} - u_{\text{true}}\|_2 / \|u_{\text{true}}\|_2$, evaluated on a fine test grid against high-fidelity analytical or numerical ground truth solutions. 

To test our two-part hypothesis, we perform two analyses. First, we fit the univariate scaling $\text{error} \approx A N^{-\alpha}$ at each fixed $\kappa$ value to analyze the function $\alpha(\kappa)$ (results in Tab.~\ref{tab:alpha_vs_kappa}). Second, we fit and compare several multivariate scaling laws, including the 
simple separable model $\text{error} \approx A N^{-\alpha} \kappa^{\gamma}$ (Tab.~\ref{tab:multivariate_fit}) and a more complex, non-separable interaction model (see Appendix~\ref{sec:alt_scaling_laws}), to quantify the coupling between width and nonlinearity.

\section{Results and Discussion}

Our experiments reveal a significant and often counterintuitive relationship between network width, PDE nonlinearity, and final solution error. We find that the practical performance of SLN-PINNs is consistently dominated by optimization challenges, which prevent the realization of theoretical approximation benefits and can lead to pathological scaling behavior where wider networks perform worse.

\subsection{Evidence of Width Scaling Pathology on a Linear Benchmark}

We first test our framework on the linear Poisson equation, whose solution with a smooth forcing term is analytic. For such highly smooth functions, classical approximation theory suggests that neural networks can achieve faster convergence rates than the general Barron rate for less regular functions.~\cite{yarotsky2017error, devore2020neural} These theoretical results suggest an error decay of $\mathcal{O}(N^{-k/d})$, where $k$ relates to the function's smoothness, which for our one-dimensional ($d=1$) problem, could imply a width scaling exponent $\alpha > 0.5$.

\begin{figure}[h]
    \centering
    \includegraphics[width=0.45\linewidth]{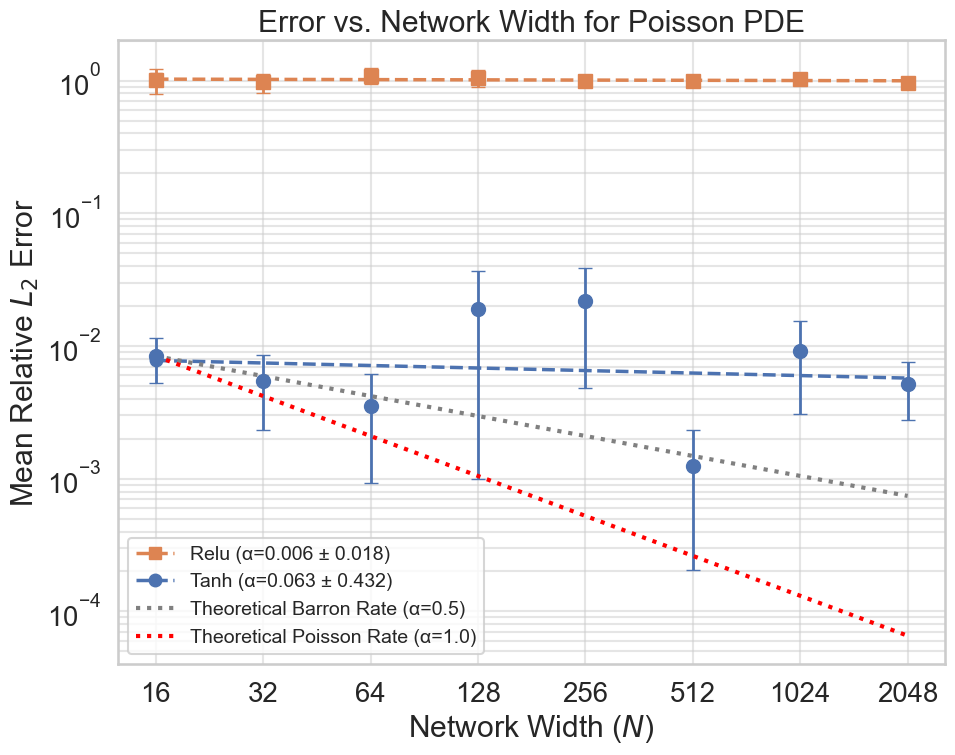}
    \caption{Error vs. Network Width ($N$) for the Poisson PDE. \texttt{Tanh} networks find low-error solutions but exhibit high variance and no clear scaling ($\alpha \approx 0.06 \pm 0.4$). \texttt{ReLU} networks fail to learn ($\alpha \approx 0.01 \pm 0.01$). The gray and red lines give the theoretical error decay rates of $\mathcal{O}(N^{-1/2})$ and $\mathcal{O}(N^{-1})$ respectively. It should be noted that the confidence intervals of the \texttt{tanh} error often intersect with theoretical estimates, but no consistent trend is observed.}
    \label{fig:poisson}
\end{figure}

The results are given in Fig.~\ref{fig:poisson}. \texttt{Tanh}-activated networks successfully converge to low-error solutions ($\approx 10^{-3}$), though with high variance across seeds and no consistent scaling trend ($\alpha \approx 0.06$). In contrast, \texttt{ReLU}-activated networks catastrophically fail to learn; the error remains high ($\approx 1.0$) regardless of network width, yielding an effective scaling exponent of $\alpha \approx 0.01$. This is a classic example of spectral bias. While a \texttt{ReLU} network can theoretically approximate the solution, its second derivative (penalized by the $u_{xx}$ term) is a sparse set of Dirac delta functions.~\cite{rahaman2019spectral} This makes it particularly ill-suited for representing the smooth, continuous derivatives required by the PDE loss~\cite{wang2022understanding}. The $C^\infty$ \texttt{tanh} activation, being smoother, suffers less severely but is still affected by a baseline failure.

\subsection{Compounding Pathology: Non-Separable Scaling with Nonlinearity}

Our central finding is that for nonlinear PDEs, the optimization failure is compounded by the problem's nonlinearity. This failure is not an artifact of averaging across different hardness levels. As shown in detail in Appendix~\ref{sec:scaling-law-tables}, fitting the scaling law $\text{error} \approx A \cdot N^{-\alpha}$ individually for each fixed $\kappa$ consistently yields width exponents $\alpha$ that are near-zero or negative. This confirms the baseline pathology is pervasive across all conditions tested.

To quantify the average effect of both width and hardness, we fit the multivariate separable power law. The results are presented in Tab.~\ref{tab:multivariate_fit}. The width exponent $\alpha$ is consistently near-zero or negative, confirming that wider networks do not help and can even be harmful. The hardness exponent $\gamma$ is generally positive, confirming that problems become harder as $\kappa$ increases, with the notable exception of the Allen-Cahn equation, which we discuss later.

\begin{table}[b]
\scriptsize
\caption{Overall scaling exponents from the separable power-law regression $\text{error} \approx A \cdot N^{-\alpha} \cdot \kappa^{\gamma}$. Here, $\alpha$ represents the average width scaling, while $\gamma$ quantifies the error's sensitivity to the hardness parameter $\kappa$. Statistical significance based on the p-value of the fitted coefficient is indicated by: $p<0.01$ (**), $p<0.001$ (***).}
\label{tab:multivariate_fit}
\centering
\sisetup{
    round-mode=places,
    round-precision=2,
    separate-uncertainty=true,
    table-align-text-post=false
}
\begin{tabular}{l l S S S S}
\toprule
\textbf{PDE} & \textbf{Activation} & {\textbf{Width Exp. ($\alpha \pm 95\%$ CI)}} & {\textbf{Hardness Exp. ($\gamma \pm 95\%$ CI)}} & {\textbf{log(A)}} & {\textbf{Adj. $R^2$}} \\
\midrule
\multirow{2}{*}{\textbf{KdV}} 
 & ReLU & -0.05 \pm 0.03 ** & 0.17 \pm 0.04 *** & -0.66 & 0.65 \\
 & Tanh & 0.00 \pm 0.01 & 0.18 \pm 0.01 *** & -0.27 & 0.51 \\
\midrule
\multirow{2}{*}{\textbf{Sine-Gordon}} 
 & ReLU & -0.32 \pm 0.07 *** & 0.28 \pm 0.08 *** & -3.40 & 0.70 \\
 & Tanh & -0.14 \pm 0.34 & 1.51 \pm 0.39 *** & -7.30 & 0.52 \\
\midrule
\multirow{2}{*}{\textbf{Allen-Cahn}} 
 & ReLU & -0.37 \pm 0.08 *** & -0.44 \pm 0.10 *** & -3.36 & 0.74 \\
 & Tanh & -0.02 \pm 0.11 & -0.03 \pm 0.12 & -6.78 & -0.03 \\
\bottomrule
\end{tabular}
\end{table}

\begin{figure}[t]
    \centering
    \includegraphics[width=\linewidth]{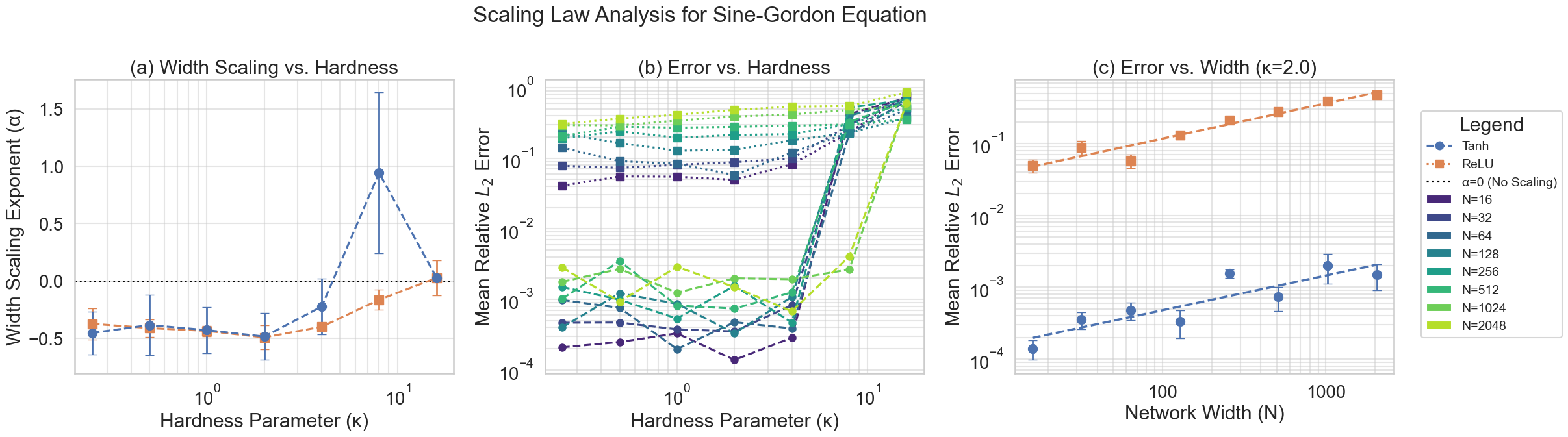}
    \caption{Scaling law analysis for the Sine-Gordon equation. \textbf{(a)} Width scaling exponent $\alpha$ vs. hardness $\kappa$. Often $\alpha < 0$, implying increasing network width also increases error. \textbf{(b)} Final error vs. hardness $\kappa$ for different network widths $N$. The final error degrades significantly at a certain inflection point of hardness, indicating some kind of regime shift. \textbf{(c)} A representative example of error vs. width $N$ for the median hardness value $\kappa=2.0$ demonstrates that increased width fails to reduce error. Error bars denote standard error over 5 seeds.}
    \label{fig:sinegordon}
\end{figure}

Fig.~\ref{fig:sinegordon} visualizes how this pathology is exacerbated as nonlinearity increases, using the Sine-Gordon equation as a representative example. We observe that the width scaling exponent $\alpha$ becomes a more complex, non-monotonic function of the hardness $\kappa$ (Fig.~\ref{fig:sinegordon}a), demonstrating that a single scaling exponent is insufficient. While the final error generally increases with hardness, the relationship between network width and error is unstable and often detrimental. Notice, furthermore, that while a change in $N$ typically results in less than an order of magnitude difference in error, a change in hardness $\kappa$ can alter the error by several orders of magnitude (Fig.~\ref{fig:sinegordon}b). This suggests that for these problems, nonlinearity can be a far more dominant factor than network width.

Nonlinear terms in the PDE create a fundamentally more complex and non-convex loss landscape. As the problem becomes more nonlinear, the optimization fails not because of a lack of network capacity but because standard gradient-based methods find it more difficult to locate a good solution. This aligns with recent theoretical work showing that standard intuitions from infinitely-wide networks (like the NTK) are misleading in these nonlinear regimes.~\cite{bonfanti2023challenges}

This complex behavior, where the effect of width appears dependent on the level of nonlinearity, strongly suggests that the simple separable scaling law is insufficient. We test this hypothesis directly by fitting a more complex, non-separable model with an interaction term. The full analysis, presented in Appendix~\ref{sec:alt_scaling_laws}, quantitatively confirms this suspicion and reveals a key mechanistic difference between the activation functions, providing strong evidence for our second hypothesis. The interaction term is statistically significant in every case for \texttt{ReLU} but in no case for \texttt{tanh}, suggesting an inheritance of $\kappa$-dependent stiffness in the former case. 

We observe similar scaling pathologies for the KdV and Allen-Cahn equations albeit with their own subtleties (see Tab.~\ref{tab:multivariate_fit}). While the dispersive (KdV) and hyperbolic (Sine-Gordon) equations show the expected trend of error increasing with hardness ($\gamma > 0$), the reactive-parabolic Allen-Cahn equation is a notable exception. For \texttt{ReLU}, $\gamma$ is significantly negative, suggesting a qualitatively different failure mechanism. For \texttt{tanh}, the model fits poorly (negative Adj. $R^2$), and it can be visually observed in Appendix~\ref{sec:other-eqs} that the error is lower but skewed by the final $N=1024$ point.

\section{Conclusion and Future Work}

In this work, we introduced a methodology for establishing empirical scaling laws for SLN-PINNs, and identified a dual optimization failure from non-convexity and spectral bias.

First, we demonstrated a width scaling pathology: standard, gradient-trained PINNs fail to achieve the positive scaling with width ($\alpha \approx 0.5$) conjectured by approximation theory. We find $\alpha \approx 0$ (wider does nothing) or $\alpha < 0$ (wider is worse), a clear sign that optimization, not approximation capacity, is the bottleneck. Second, we identified a compounding effect of nonlinearity, where increasing nonlinearity $\kappa$ (and, for our set of PDEs, solution frequency) causes the scaling law itself to break down. We showed (Appendix~\ref{sec:alt_scaling_laws}) that for \texttt{ReLU}, this manifests as a non-separable interaction, while for \texttt{tanh}, network width ceases to be a statistically significant factor at all. We further argue that this effect depends heavily on the specific PDE.

This study provides a quantitative benchmark on the practical limits of standard PINNs. It suggests that a ``brute-force'' approach of simply using wider shallow networks is an inefficient strategy. We wish to emphasize that the main purpose of this work is a call-to-action for such scaling studies in other settings. This work uses Adam, SLN-PINNs, and a small set of PDEs with one spatial dimension to make the benefits of this approach clear. Future work should aim to benchmark and identify architectures (e.g., multi-layer, Fourier features, attention-based) and optimizers (e.g., adaptive weighting, second-order methods) that can (1) close the gap between theoretical and empirical scaling of width and (2) demonstrate robustness to problem hardness and spectral bias. Overall, while it is reasonable to assume that nonlinearity causes difficulty for PINNs, it should be further investigated whether different types of nonlinear equations have different scaling laws.

\newpage
\bibliographystyle{unsrtnat}
\bibliography{bibliography}

@article{raissi2019physics,
  title={Physics-informed neural networks: A deep learning framework for solving forward and inverse problems involving nonlinear partial differential equations},
  author={Raissi, Maziar and Perdikaris, Paris and Karniadakis, George E},
  journal={Journal of Computational Physics},
  volume={378},
  pages={686--707},
  year={2019},
  publisher={Elsevier}
}

@article{krishnapriyan2021characterizing,
  title={Characterizing possible failure modes in physics-informed neural networks},
  author={Krishnapriyan, Aditi S and Gholami, Amir and McQueen, Dillon and Lapedriza, Agueda and Tort-Cases, Oxana and Buhler, Julius and Mahoney, Michael W},
  journal={Advances in Neural Information Processing Systems},
  volume={34},
  year={2021}
}

@article{barron1993universal,
  title={Universal approximation bounds for superpositions of a sigmoidal function},
  author={Barron, Andrew R},
  journal={IEEE Transactions on Information Theory},
  volume={39},
  number={3},
  year={1993},
  publisher={IEEE}
}

@inproceedings{reyes2021learning,
  title={Learning physics-based models of fluid dynamics from sparse data},
  author={Reyes, Bryan and Howard, Amanda A and Perdikaris, Paris and Karni, S},
  booktitle={AAAI Conference on Artificial Intelligence},
  volume={35},
  year={2021}
}

@article{arzani2021uncovering,
  title={Uncovering near-wall blood flow from sparse data with physics-informed neural networks},
  author={Arzani, Amir and Wang, Jian-Xun and D'Souza, Roshan M},
  journal={Physics of Fluids},
  volume={33},
  number={7},
  year={2021},
  publisher={AIP Publishing}
}

@inproceedings{bonfanti2023challenges,
  title={The Challenges of the Nonlinear Regime for Physics-Informed Neural Networks},
  author={Bonfanti, Andrea and Bruno, Giacomo and Cipriani, Carlo},
  booktitle={Advances in Neural Information Processing Systems},
  volume={36},
  year={2023}
}

@article{cybenko1989approximation,
  title={Approximation by superpositions of a sigmoidal function},
  author={Cybenko, George},
  journal={Mathematics of Control, Signals and Systems},
  volume={2},
  number={4},
  year={1989},
  publisher={Springer}
}

@article{hornik1989multilayer,
  title={Multilayer feedforward networks are universal approximators},
  author={Hornik, Kurt and Stinchcombe, Maxwell and White, Halbert},
  journal={Neural Networks},
  volume={2},
  number={5},
  year={1989},
  publisher={Elsevier}
}

@article{kaplan2020scaling,
  title={Scaling laws for neural language models},
  author={Kaplan, Jared and McCandlish, Sam and Henighan, Tom and Brown, Tom B and Chess, Benjamin and Child, Rewon and Gray, Scott and Radford, Alec and Wu, Jeffrey and Amodei, Dario},
  journal={arXiv preprint arXiv:2001.08361},
  year={2020}
}

@inproceedings{jacot2018neural,
  title={Neural tangent kernel: Convergence and generalization in neural networks},
  author={Jacot, Arthur and Gabriel, Franck and Hongler, Cl{\'e}ment},
  booktitle={Advances in Neural Information Processing Systems},
  volume={31},
  year={2018}
}

@article{wang2022understanding,
  title={Understanding and mitigating gradient flow pathologies in physics-informed neural networks},
  author={Wang, Sifan and Yu, Xinling and Perdikaris, Paris},
  journal={SIAM Journal on Scientific Computing},
  volume={44},
  number={1},
  year={2022},
  publisher={SIAM}
}

@article{moseley2023finite,
  title={Finite basis physics-informed neural networks (FBPINNs): a scalable domain decomposition approach for solving differential equations},
  author={Moseley, Ben and Markham, Andrew and Nissen-Meyer, Tarje},
  journal={Advances in Computational Mathematics},
  volume={49},
  number={4},
  year={2023},
  publisher={Springer}
}

@article{you2018effect,
  title={The effect of network width on the performance of large-batch training},
  author={You, Yang and Gitman, Igor and Ginsburg, Boris},
  journal={arXiv preprint arXiv:1812.04252},
  year={2018}
}

@article{yarotsky2017error,
  title={Error bounds for approximations with deep {ReLU} networks},
  author={Yarotsky, Dmitry},
  journal={Neural Networks},
  volume={94},
  year={2017},
  publisher={Elsevier}
}

@article{devore2020neural,
  title={Neural network approximation},
  author={DeVore, Ronald A and Hanin, Boris and Petrova, Guergana},
  journal={Acta Numerica},
  volume={30},
  year={2021},
  publisher={Cambridge University Press}
}

@inproceedings{rahaman2019spectral,
  title={On the Spectral Bias of Neural Networks},
  author={Rahaman, Nasim and Baratin, Aristide and Arpit, Devansh and Draxler, Felix and Lin, Min and Hamprecht, Fred A. and Bengio, Yoshua and Courville, Aaron},
  booktitle={36th International Conference on Machine Learning},
  year={2019},
  organization={PMLR},
  volume={97},
}

%%%%%%%%%%%%%%%%%%%%%%%%%%%%%%%%%%%%%%%%%%%%%%%%%%%%%%%%%%%%

\newpage
\appendix

\section{Allen-Cahn and KdV Scaling Analysis}
\label{sec:other-eqs}

In this section, we present the full scaling law analysis for the Allen-Cahn and KdV equations. 

\begin{figure}[ht]
    \centering
    \includegraphics[width=\linewidth]{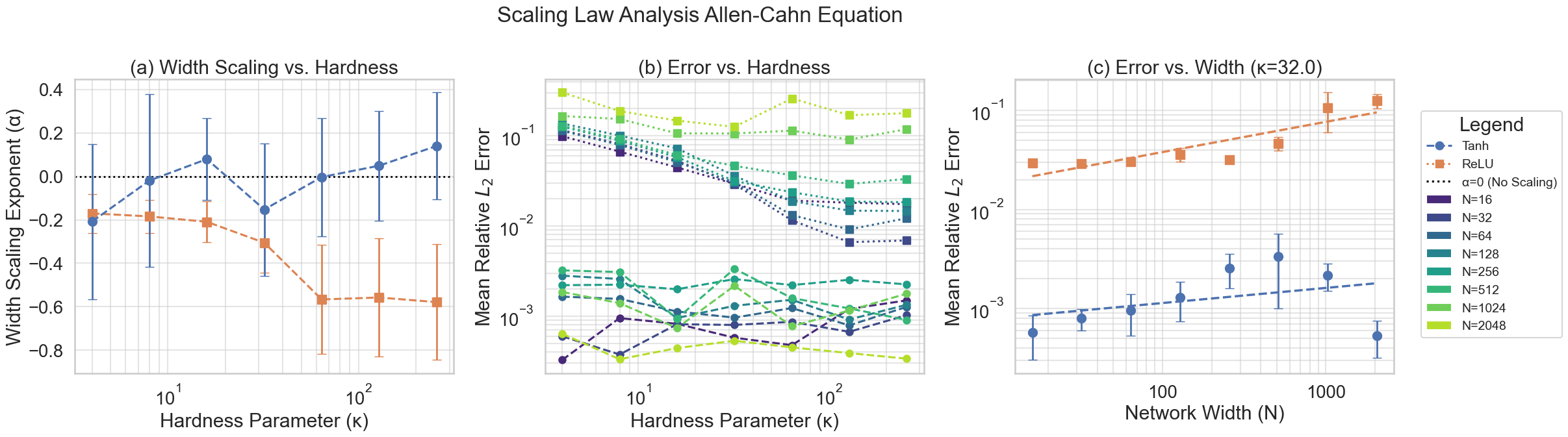}
    \caption{Scaling law analysis for the Allen-Cahn equation. \textbf{(a)} Width scaling exponent $\alpha$ vs. hardness $\kappa = 1/D$. For both activations, $\alpha$ is consistently negative, indicating wider networks perform worse. \textbf{(b)} Final error vs. hardness $\kappa$. \texttt{Tanh} networks (dashed lines) achieve low error that is remarkably stable against increasing hardness, while \texttt{ReLU} networks (dotted lines) perform orders of magnitude worse. \textbf{(c)} A representative example of error vs. width $N$ for the median hardness value $\kappa=32.0$, clearly showing the negative scaling trend ($\alpha < 0$). Error bars denote standard error over 5 seeds.}
    \label{fig:allencahn}
\end{figure}

Fig.~\ref{fig:allencahn} shows the scaling analysis for the Allen-Cahn equation. For both \texttt{tanh} and \texttt{ReLU} activations, the width scaling exponent $\alpha$ is consistently negative or near-zero, indicating that increasing network width provides no benefit and often harms performance. This pathology is especially severe for \texttt{ReLU} networks, where the scaling exponent becomes progressively more negative as the problem hardness $\kappa$ increases (Fig.~\ref{fig:allencahn}a). Furthermore, there is a performance gap between the activation functions; \texttt{tanh} networks achieve a low error that remains remarkably stable across all hardness values, whereas \texttt{ReLU} networks perform orders of magnitude worse with a width scaling exponent that becomes more negative as hardness increases.

\begin{figure}[ht]
    \centering
    \includegraphics[width=\linewidth]{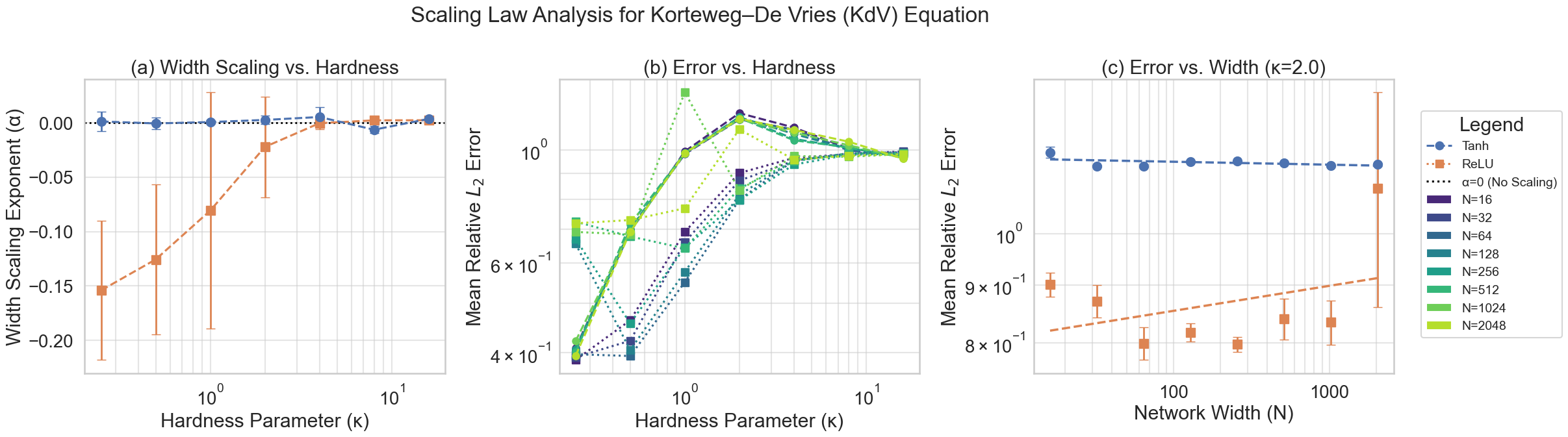} 
    \caption{Scaling law analysis for the Korteweg-de Vries (KdV) equation. \textbf{(a)} Width scaling exponent $\alpha$ vs. hardness $\kappa = A$. For both \texttt{ReLU} and \texttt{tanh}, $\alpha$ is consistently near or below zero, indicating no performance gain from increasing network width. \textbf{(b)} Final error vs. hardness $\kappa$. Error increases with hardness for both activations, though \texttt{tanh} generally performs better. \textbf{(c)} A representative example of error vs. width $N$ for $\kappa=2.0$, visualizing the flat ($\alpha \approx 0$) or negative scaling. Error bars denote standard error over 5 seeds.}
    \label{fig:kdv}
\end{figure}

Fig.~\ref{fig:kdv} shows the scaling analysis for the KdV equation. A sharp degradation in solution quality is observed as the hardness parameter $\kappa$ increases (Fig.~\ref{fig:kdv}b). The analysis of the width scaling exponent $\alpha$ again reveals a failure to benefit from larger networks. For \texttt{tanh}, $\alpha$ remains consistently close to zero across all hardness values, indicating no performance gain from increasing width. For \texttt{ReLU}, $\alpha$ begins as negative and trends toward zero as $\kappa$ increases (Fig.~\ref{fig:kdv}a). While \texttt{tanh} generally achieves a lower error than \texttt{ReLU}, their overall performance is more comparable than in the Allen-Cahn case, suggesting some interaction between activation function and the type of PDE nonlinearity.

\newpage
\section{Table of Univariate Regressions on Width}
\label{sec:scaling-law-tables}

\begin{table}[ht]
\caption{Fitted width scaling exponents $\alpha$ for each hardness parameter $\kappa$ and activation function. The exponent is derived from fitting the mean error over 5 seeds to the power law $\text{error} \approx A \cdot N^{-\alpha}$.}
\label{tab:alpha_vs_kappa}
\centering
\sisetup{
    round-mode=places,
    round-precision=2,
    separate-uncertainty=true,
    table-align-text-post=false
}
\begin{tabular}{l S[table-format=4.2] S[table-format=-1.2(2)] S[table-format=-1.2(2)]}
\toprule
\textbf{PDE} & {\textbf{Hardness ($\kappa$)}} & {\textbf{ReLU ($\alpha \pm 95\%$ CI)}} & {\textbf{Tanh ($\alpha \pm 95\%$ CI)}} \\
\midrule
\multirow{1}{*}{\textbf{Poisson}} & N/A & 0.01 \pm 0.02 & 0.06 \pm 0.43 \\
\midrule
\multirow{7}{*}{\textbf{KdV}} 
 & 0.25 & -0.15 \pm 0.06 & 0.00 \pm 0.01 \\
 & 0.50 & -0.13 \pm 0.07 & 0.00 \pm 0.01 \\
 & 1.00 & -0.08 \pm 0.11 & 0.00 \pm 0.01 \\
 & 2.00 & -0.02 \pm 0.05 & 0.00 \pm 0.01 \\
 & 4.00 & 0.00 \pm 0.01 & 0.01 \pm 0.01 \\
 & 8.00 & 0.00 \pm 0.01 & -0.01 \pm 0.01 \\
 & 16.00 & 0.00 \pm 0.01 & 0.00 \pm 0.01 \\
\midrule
\multirow{7}{*}{\textbf{Sine-Gordon}} 
 & 0.25 & -0.37 \pm 0.14 & -0.45 \pm 0.19 \\
 & 0.50 & -0.41 \pm 0.08 & -0.38 \pm 0.26 \\
 & 1.00 & -0.43 \pm 0.04 & -0.43 \pm 0.20 \\
 & 2.00 & -0.49 \pm 0.10 & -0.48 \pm 0.20 \\
 & 4.00 & -0.40 \pm 0.03 & -0.22 \pm 0.24 \\
 & 8.00 & -0.16 \pm 0.09 & 0.94 \pm 0.70 \\
 & 16.00 & 0.03 \pm 0.16 & 0.03 \pm 0.01 \\
\midrule
\multirow{7}{*}{\textbf{Allen-Cahn}} 
 & 4.0 & -0.17 \pm 0.09 & -0.21 \pm 0.36 \\
 & 8.0 & -0.18 \pm 0.08 & -0.02 \pm 0.40 \\
 & 16.0 & -0.21 \pm 0.09 & 0.08 \pm 0.19 \\
 & 32.0 & -0.31 \pm 0.14 & -0.15 \pm 0.30 \\
 & 64.0 & -0.57 \pm 0.25 & 0.00 \pm 0.27 \\
 & 128.0 & -0.56 \pm 0.27 & 0.05 \pm 0.25 \\
 & 256.0 & -0.58 \pm 0.27 & 0.14 \pm 0.25 \\
\bottomrule
\end{tabular}
\end{table}

\clearpage
\section{Scaling Laws for Interaction Between Nonlinearity and Width}
\label{sec:alt_scaling_laws}

To test our hypothesis that the simple, separable scaling law (Tab.~\ref{tab:multivariate_fit}) is insufficient, we fit a more complex, non-separable interaction model. Previously the simple model was $\log(\text{error}) = \beta_0 + \beta_N \log(N) + \beta_\kappa \log(\kappa)$. Tab.~\ref{tab:interaction_model_fit} gives the results for fitting a model with some interaction term, which is now defined as:
\begin{equation}
    \label{eq:reg-interaction-model}
    \log(\text{error}) = \beta_0 + \beta_N \log(N) + \beta_\kappa \log(\kappa) + \nu (\log(N)\cdot\log(\kappa)).
\end{equation}
A statistically significant interaction coefficient $\nu$ (i.e., $p < 0.05$) would suggest that the scaling with width $N$ is dependent on the nonlinearity $\kappa$.

\begin{table}[ht]
\scriptsize
\caption{Scaling exponents for the interaction model defined in~\eqref{eq:reg-interaction-model}. Statistical significance based on the p-value of the fitted coefficient is indicated by: $p<0.01$ (**), $p<0.001$ (***).}
\label{tab:interaction_model_fit}
\centering
\sisetup{
    round-mode=places,
    round-precision=3,
    table-align-text-post=false
}
\begin{tabular}{l l S S S S S}
\toprule
\textbf{PDE} & \textbf{Activation} & {\textbf{$\beta_0$ (Intercept)}} & {\textbf{$\beta_N$ (Width)}} & {\textbf{$\beta_\kappa$ (Hardness)}} & {\textbf{$\nu$ (Interaction)}} & {\textbf{Adj. $R^2$}} \\
\midrule
\multirow{2}{*}{\textbf{KdV}} 
& ReLU & -0.809 *** & 0.083 *** & 0.386 *** & \textbf{-0.042 ***} & \textbf{0.741} \\
& Tanh & -0.271 ** & -0.001 & 0.178 ** & 0.000 & 0.498 \\
\midrule
\multirow{2}{*}{\textbf{Sine-Gordon}} 
& ReLU & -3.716 *** & 0.382 *** & 0.740 *** & \textbf{-0.089 ***} & \textbf{0.758} \\
& Tanh & -8.094 *** & 0.296 & 2.666 *** & -0.222 & 0.542 \\
\midrule
\multirow{2}{*}{\textbf{Allen-Cahn}} 
& ReLU & -1.202 ** & -0.048 & -1.061 *** & \textbf{0.120 ***} & \textbf{0.811} \\
& Tanh & -7.800 *** & 0.212 & 0.260 & -0.057 & -0.008 \\
\bottomrule
\end{tabular}
\end{table}

The results in Tab.~\ref{tab:interaction_model_fit} provide strong evidence for a non-separable scaling relationship, particularly for \texttt{ReLU} networks. For every nonlinear PDE, the interaction term $\nu$ for \texttt{ReLU} is statistically significant, and the adjusted $R^2$ of the interaction model is notably higher than that of the simple separable model in Table~\ref{tab:multivariate_fit}. This is the statistical signature of a brittle failure mode: the scaling of \texttt{ReLU} with network width is actively exacerbated by problem hardness. This is consistent with the spectral bias theory, as the non-smooth ReLU activation creates a particularly difficult optimization landscape when forced to represent the high-frequency solutions associated with higher $\kappa$~\cite{rahaman2019spectral}.

In contrast, both the width term $\beta_N$ and the interaction term $\nu$ for \texttt{tanh} are not statistically significant for any of the PDEs. For \texttt{tanh} networks, network width appears to be an irrelevant variable for predicting the final error. The failure mode is not one of complex interaction, but rather a consistent failure to benefit from increased capacity. This points to a more ``graceful'' failure mode. The inherent smoothness of the \texttt{tanh} function, while still subject to spectral bias, leads to a more stable, albeit suboptimal, optimization dynamic.

The Allen-Cahn equation serves as a particularly interesting case study that highlights the limits of this power-law model. For the \texttt{tanh} network, the extremely poor fit of the model (negative adjusted $R^2$) is itself a result; it confirms the visual evidence from Fig.~\ref{fig:allencahn}, where the error is low and stable across all $N$ and $\kappa$. For the \texttt{ReLU} network on Allen-Cahn, we note the unusual negative, statistically-significant hardness coefficient $\beta_{\kappa}$, a behavior not seen in the other PDEs that suggests its failure mechanism may be qualitatively different in this reactive-parabolic setting.

\end{document}